\documentclass[conference]{IEEEtran}
\IEEEoverridecommandlockouts
\usepackage{cite}
\usepackage{amsmath,amssymb,amsfonts, bbm}
\usepackage{algorithmic}
\usepackage{graphicx}
\usepackage{textcomp}
\usepackage{xcolor}
\def\BibTeX{{\rm B\kern-.05em{\sc i\kern-.025em b}\kern-.08em
    T\kern-.1667em\lower.7ex\hbox{E}\kern-.125emX}}
\usepackage[marginal]{footmisc}
\begin{document}

\title{Cost-Effective Active Labeling for Data-Efficient Cervical Cell Classification\vspace{-9pt}
\thanks{Corresponding author: Youyi Song.\\
This work is partly supported by a NSFC grant: 62406343.}
}
\author{\IEEEauthorblockN{Yuanlin Liu}
\IEEEauthorblockA{\textit{School of Science} \\
\textit{China Pharmaceutical University}\\
Nanjing, China\\
liuyuanlin@stu.cpu.edu.cn}\vspace{4pt}
\IEEEauthorblockN{Youyi Song}
\IEEEauthorblockA{\textit{School of Science} \\
\textit{China Pharmaceutical University}\\
Nanjing, China \\
youyisong@cpu.edu.cn}\vspace{-22pt}
\and
\IEEEauthorblockN{Zhihan Zhou}
\IEEEauthorblockA{\textit{School of Science} \\
\textit{China Pharmaceutical University}\\
Nanjing, China \\
zhouzhihan@stu.cpu.edu.cn}\vspace{4pt}
\IEEEauthorblockN{Mingqiang Wei}
\IEEEauthorblockA{\textit{College of Computer Science and Technology} \\
\textit{Nanjing University of Aeronautics and Astronautics}\\
Nanjing, China \\
mqwei@nuaa.edu.cn}\vspace{-22pt}
}

\maketitle

\begin{abstract}
Information on the number and category of cervical cells is crucial for the diagnosis of cervical cancer.
However, existing classification methods capable of automatically measuring this information require the training dataset to be representative, which consumes an expensive or even unaffordable human cost.
We herein propose active labeling that enables us to construct a representative training dataset using a much smaller human cost for data-efficient cervical cell classification.
This cost-effective method efficiently leverages the classifier's uncertainty on the unlabeled cervical cell images to accurately select images that are most beneficial to label.
With a fast estimation of the uncertainty, this new algorithm exhibits its validity and effectiveness in enhancing the representative ability of the constructed training dataset.
The extensive empirical results confirm its efficacy again in navigating the usage of human cost, opening the avenue for data-efficient cervical cell classification.

\end{abstract}


\section{Introduction}

It remains a great challenge for cervical cancer prevention worldwide~\cite{alfaro2021removing}.
Cervical cancer screening based on microscopic cytology has been widely recognized as a feasible and effective solution to a high-quality prevention~\cite{perkins2023cervical}.
A crucial step in such a screening is to accurately classify cervical cells in the microscope slide into their cytological categories, from which one can readily measure the abnormality extent of the cervix and estimate its progression for making the prevention decision appropriately and in a timely manner~\cite{song2024cell}.

\begin{figure}[t]
  \centering
  \includegraphics[width = 3.4in, height = 1.7in]{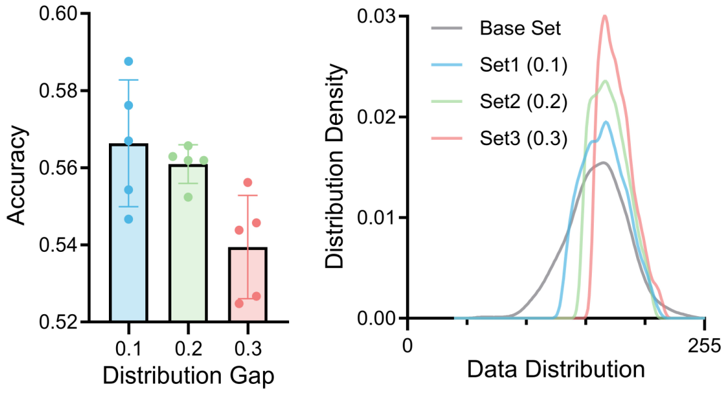}
  \caption{The illustration of how the representativeness of the training dataset affects the classification accuracy of cervical cell classification.
  %
  %We employ the distribution gap to measure the representativeness between the training dataset and test dataset, and the right plot shows an example of how we calculate it: for each dataset, we use the mean of RGB values of the cell image to mimic the data distribution, and the gap is measured by the total area of the difference between distributions.
  %
  }
  \label{FigProblem}
  \vspace{-6pt}
\end{figure}

Advanced deep learning techniques along with the enhanced computational capacity powered by GPU cards have been theoretically and empirically showed being able to handle cervical cell classification task~\cite{bengio2021deep,menghani2023efficient}.
However, their success heavily depends on the quality of training dataset that is required to be composed of independent and identically distributed samples and representative enough to ensure the trained classifier performing similarly well to all unseen data~\cite{esteva2019guide}; see Fig.~\ref{FigProblem} about how the representativeness of the training dataset affects the learning performance of cervical cell classification for example.
As each microscope slide contains over several million cells and unseen cells cover several billion slides, a proper training dataset collected by the routine way will be formidably large and consume an expensive or even unaffordable human cost~\cite{von2021democratising}.

Many efforts have been made to reduce the human cost in class label collection by leveraging classification rule information from other sources of training dataset~\cite{chen2023vlp,oza2023unsupervised} (\emph{e.g.} publicly available natural image sets) or unlabeled cervical cell images via transfer learning~\cite{niu2021decade} and semi-supervised learning~\cite{yang2022survey}, just to mention a few.
These approaches are indeed effective on reducing the human cost, but all ignore the central problem: \emph{how to maximally utilize an allowed human cost to construct a representative training dataset?}
We here argue that if this problem can be properly solved, then it becomes possible to construct a representative training dataset under the controlled human cost and we can further improve the classification performance by combining with these effective techniques.

\begin{figure*}[t]
   \centering
   \includegraphics [width = 7.0in, height = 2.38in]{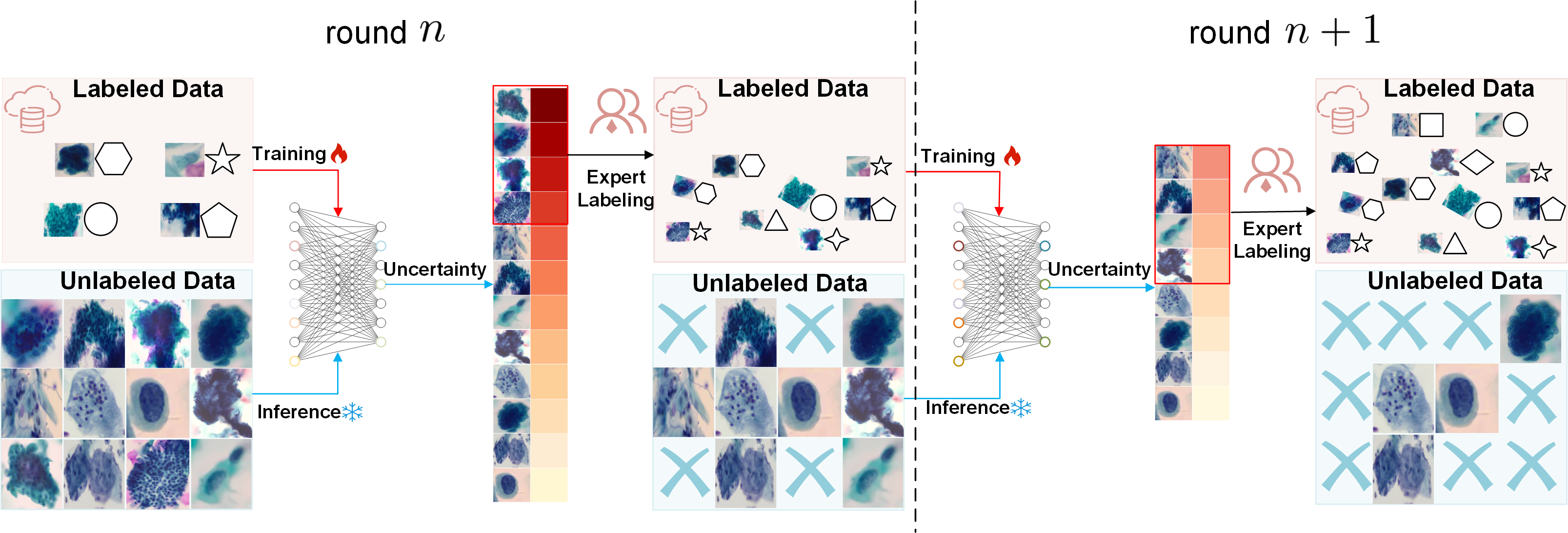}
   \caption{The illustration of the proposed active labeling algorithm that enables to construct a representative training dataset for cervical cell classification.   
  %
  %This cost-effective method simultaneously constructs the training dataset and trains the deep classifier.
  %
  %At each labeling selection round, after training the classifier, it selects cervical cell image for labeling according to the uncertainty of the classifier on the all unlabeled images until the allowed labeling cost has been exhausted; the images with the highest uncertainty value are selected.
  }
\label{Framework}
\end{figure*}

We herein propose active labeling, an automatic data labeling framework capable of maximally exploiting the human cost of constructing a representative training dataset for data-efficient cervical cell classification.
The key idea is to examine the classifier's uncertainty on the unlabeled cervical cell images for harnessing the information about which images are most beneficial to label in terms of enhancing the representative ability of the constructing training dataset.
We yet develop a tailored algorithm to quickly estimate the uncertainty, which improves the selection quality of images for labeling, further efficiently navigating the usage of human cost. 
We conducted a variety of experiments to evaluate the validity and found that the proposed method requires a much smaller human cost to construct a comparably representative training dataset.
%
%Put differently, this means that it constructs a far more representative training dataset under the same human cost given, being cost-effective and data-efficient for cervical cell classification.   
%
%We finally hope that this cost-effective labeling approach could be a useful tool in decision making about cervical cancer prevention.

%\input{RelatedWorks}

\section{Methodology}

\subsection{Problem Description}

Given $N$ unlabeled cervical cell images $\mathbf{x}_1, \cdots, \mathbf{x}_N$ and a deep cell classification model $f\in\mathcal{F}:\mathcal{X}\to\mathcal{Y}$, where $\mathcal{F}$, $\mathcal{X}$ and $\mathcal{Y}$ stand for the classification function space, image space and class space, respectively, our goal is to find out a labeling indicator $\ell: \mathcal{X}\to\{0, 1\}$ where $\ell(\mathbf{x})=1$ means sending the image $\mathbf{x}$ to the human expert for querying its class label $\mathbf{y}$, such that the constructed training dataset $\{(\mathbf{x}_i, \mathbf{y}_i): \ell(\mathbf{x}_i)=1\}_{i=1}^N$ with the size $n\ll{N}$ is representative enough for the deep model per the definition as below, 
\par\nobreak\noindent
\begin{align}
    \label{Objective}
    \lim_{N\to\infty}\mathbb{E}_\mathbf{x}[(&\hat{f}^*_N(\mathbf{x})-\hat{f}^*_n(\mathbf{x}))^2]=0,\\
    \text{s.t.~~} &\hat{f}^*_N=\inf_{f\in\mathcal{F}}\sum_{i=1}^NL(f(\mathbf{x}_i),\mathbf{y}_i),\notag\\
    &\hat{f}^*_n=\inf_{f\in\mathcal{F}}\sum_{i=1}^NL(f(\mathbf{x}_i),\mathbf{y}_i|\ell(\mathbf{x}_i)=1),\notag
\end{align}
where $L$ denotes any admissible classification loss function folded by possible regularization terms.
In other words, we construct the representative training dataset via the labeling indicator $\ell$ that determines which images are most beneficial to label, in a way of increasing the representativeness of the training data under the controlled labeling cost.

\subsection{The Proposed Active Labeling Framework}

Fig.~\ref{Framework} graphically depicts the proposed solution of Eq.~\ref{Objective}.
The key idea is to leverage the uncertainty $u(\mathbf{x})$ of the deep model on the image $\mathbf{x}$ based on the perspective that the label information of images on which the model produces uncertain predictions enhances the model better. 
We here are interested in the scenario where no any pretrained model is available and yet no any label information available at the first, since it is a general setting in cervical cell classification in practice~\cite{cheng2021robust}.
Therefore, at the first round, we assign a random class label $\mathbf{y}\in\mathcal{Y}$ uniformly to each unlabeled image $\mathbf{x}_i$, and then select the first $\lfloor n/r\rfloor$ images (rounded down to the largest integer $\leq \frac{n}{r}$) with a high uncertainty value $u(\mathbf{x})$ to the human expert for querying their labels, where $r$ is the allowed round number in the whole labeling process.
At the remaining rounds, we just use the images with the class label to fine-tune the model, and perform the same image selection procedure for labeling.

Moreover, in order to speed up the selection process and enhance the estimation quality of model's certainty for a better selection, we also develop a tailored updating rule of the classification model as the following,
\par\nobreak\noindent
\begin{align}
  f^{(t+1)}=f^{(t)} -\gamma^{(t)}\nabla\bigg(\sum_{i=1}^{b}\psi\big(u(\mathbf{x}_i)\big)L\big(f^{(t)}(\mathbf{x}_i), \mathbf{y}_i\big)\bigg), 
  \label{Loss}
\end{align}
where $f^{(t)}$ and $\gamma^{(t)}$ denote the classification model and learning rate at the training iteration $t$, respectively, and $b$ represents the batch size.
The main modification is to weight the loss $L(f^{(t)}(\mathbf{x}_i), \mathbf{y}_i)$ based on the uncertainty value $u(\mathbf{x}_i)$ adjusted by a proper way of the function $\psi$, with the goal of assigning a large loss weight to images on which the model exhibits high uncertainty, thereby speeding up the training process and putting more emphasis on such images for reducing the model uncertainty which in turn helps to improve the selection quality of images for labeling.

\subsection{Choosing the Form of $u$ and $\psi$}

We finally discuss on the form of the uncertainty measure $u(\mathbf{x})$ and loss weight adjustment function $\psi$.
As we are just required to estimate the relative proportion of the uncertainty of the model on the $N$ unlabeled cell images, not the exact value, both in the image selection procedure for labeling and in the loss weighting step for classifier updating, we choose the form of them as simple as possible for maintaining the computational efficiency.

Specifically, let $[f_1(\mathbf{x}), \cdots, f_K(\mathbf{x})]^T$ be the output (normalized) of the deep classification model for a $K$-class cervical cell classification problem; here note that the normalization means $\sum_{k=1}^Kf_k(\mathbf{x})=1$.
We choose $u(\mathbf{x})=1-\max_{k\in[K]}f_k(\mathbf{x})$.
It is clear that via this form $u$ produces a very small value when the model is confident, i.e. $\max_kf_k(\mathbf{x})\approx1$ while a large value when many of $f_k(\mathbf{x})$ having a similar value, thereby being a proper measure of model's uncertainty as expected.
Likewise, we choose $\psi(u(\mathbf{x}))=\mathcal{N}_b(\alpha\frac{u(\mathbf{x})-\min_{i\in[b]}u(\mathbf{x}_i)}{\max_{i\in[b]}u(\mathbf{x}_i)-\min_{i\in[b]}u(\mathbf{x}_i)})$, where $\mathcal{N}_b$ denotes the normalization function such that the summation of the resulted $\psi(u(\mathbf{x}))$ in the data batch with the size of $b$ equals to $1$.
We can see that $\psi(u(\mathbf{x}))$ possesses a positive relationship with $u(\mathbf{x})$, though the exact vale is balanced by the hyperparameter $\alpha\in\mathbb{R}^+$, thereby being a successful realization of our idea of assigning a large loss weight to data points with a large value of model's uncertainty.

\section{Experiments}

\subsection{Datasets}

To evaluate the validity of the proposed method, we employed the \emph{HiCervix} dataset~\cite{cai2024hicervix}.
It contains $40,299$ cervical cells with $23$ cytological categories from $4,496$ microscope sides.
The $23$ categories include $8$ negative classes, $10$ positive classes, and $5$ non-cell categories.
To the best of our knowledge, it is the largest publicly available cervical cell classification dataset with the most categories.
We believe that it is likely to achieve an accurate evaluation of the proposed method in such a large dataset.

\subsection{Experimental Setup}

We resized the cell images to $384\times384$ for speeding up the training process; we empirically found that the classification performance improved not significantly with a larger size than this.
We employed ResNet$18$ as the backbone, and set the batch size as $42$ which balances well between the GPU memory consumption and classification performance.
The learning rate is initialized as $0.001$ and adjusted by the cosine reduction scheme to $0.00001$, which gives a more detailed learning~\cite{xiong2022learning}.
We employed AdamW as the optimizer~\cite{kingma2014adam}, with the weight decay parameter as (0.00001, $\beta_1$=0.9, $\beta_2$=0.999), and trained 100 epochs.
The loss function $\ell$ is specified as cross-entropy loss function which is commonly used in the cervical cell classification applications.

\subsection{Experiment Results}

\begin{figure}[t]
   \centering
   \includegraphics [width = 3.4in, height = 2.12in]{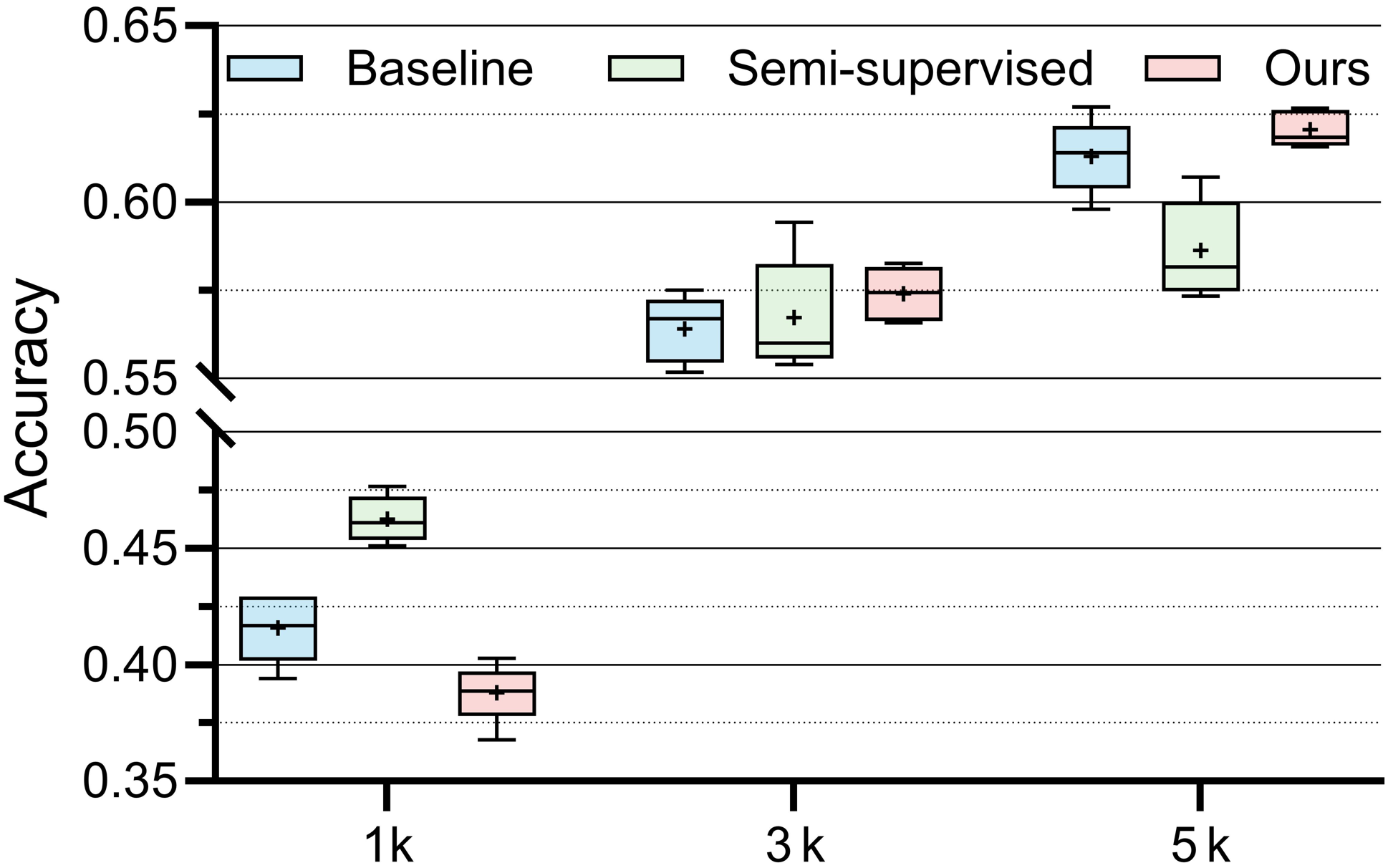}
   \caption{Cervical cell classification accuracy results ($5$ independent trials) of three different methods with the varying of the labeling cost ($1$K, $3$K and $5$K represent the number of labeled cell image pairs used in training the deep classifier).
   }
\label{FigRes}
\vspace{-6pt}
\end{figure}

We first look at how the proposed active labeling increases the representativeness of the training dataset via the lens of the improvement of the cervical cell classification accuracy. 
We compared the classification accuracy of different methods in the setting where the labeling cost varies, from $1$K to $5$K with the interval as $2$K.
We produced the results of all compared methods by $5$ independent trials and presented them in Fig.~\ref{FigRes}.

We can see that all methods produced an increasing accuracy with more labeled images used in training, but our method increases most fast, which implies the best usage of our method in the training dataset.
In other words, with more labeling cost given, our method could construct a more representative training dataset than the compared methods.
We also observed that our method achieves the smallest accuracy variance compared to other methods in all scenarios, and with the increasing of the labeling cost, the variance is generally reduced.
This finding suggests that the proposed active labeling method is more robust and the robust capability can be further improved with more labeling cost given, which implies that the construction quality of the training dataset by our method could be boosted with more cost given, and is more suitable in big data scenarios.
Putting together, it may be safe to conclude that the proposed active labeling method is able to construct a highly representative training dataset given the same amount of labeling cost, which makes its applicability in applications where the labeling cost is expensive but a large set of data is required, such as the targeting cervical cell classification application.

\begin{figure}[t]
   \centering
   \includegraphics [width = 3.4in, height = 2.77in]{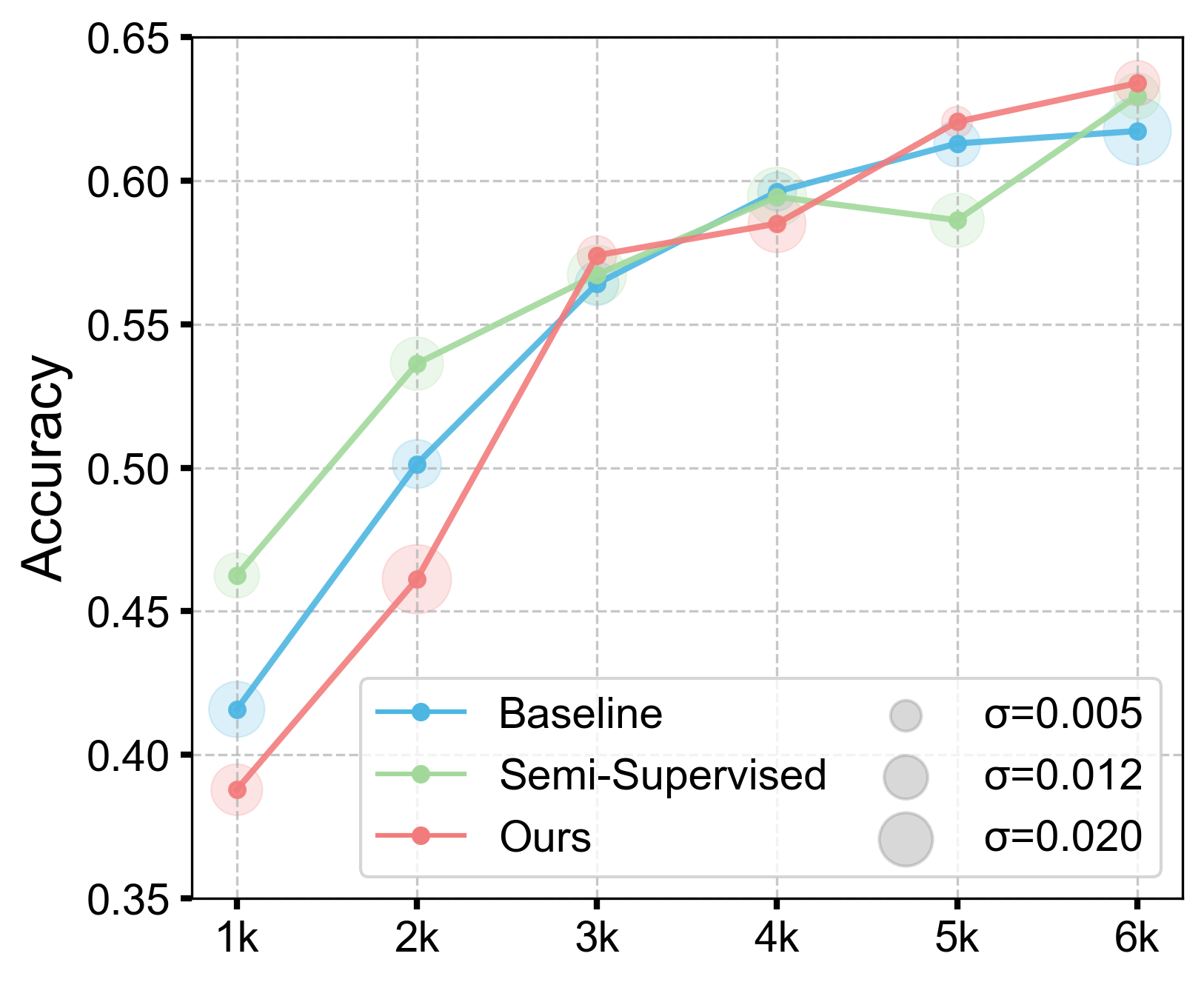}
   \caption{Cervical cell classification accuracy results ($5$ independent trials) of three different methods with more detailed labeling costs from $1$K to $6$K with the interval as $1$K; the size of the circle at each data point represents the variance of the classification accuracy on $5$ trials.}
\label{FigCost}
\vspace{-4pt}
\end{figure}

We next examine the trend of the representativeness boosted by the proposed method with more labeling cost given.
We thus conducted experiments under the more dense cost setting, from $1$K to $6$K with the interval as $1$K.
Likewise, we produced the results by $5$ independent trials, and the classification accuracy results are shown in Fig.~\ref{FigCost}.

It is clearly observed that the proposed active labeling exhibits a far more fast increasing trend of the classification accuracy with more labeling cost given.
The increasing trend is also maintained under $6$K labels given, while the `Baseline' nearly stops to increase and the `Semi-supervised' is not stable to keep an increasing trend; it even shows a decreasing trend when the labeling cost is $5$K.
The above empirical findings suggest that only our method can consistently enhance the representativeness of the constructed training dataset, and that the construction ability is more powerful when more labeling cost is given, showing its ability in the deep learning context.  
Also note that the proposed method produced the results with a decreasing variance with the increasing of the labeling cost, implying that it becomes more trustworthy when the training dataset is large enough, a desired property of deep learning in healthcare applications yet in cervical cell classification.

\begin{figure}[t]
   \centering
   \includegraphics [width = 3.4in, height = 1.62in]{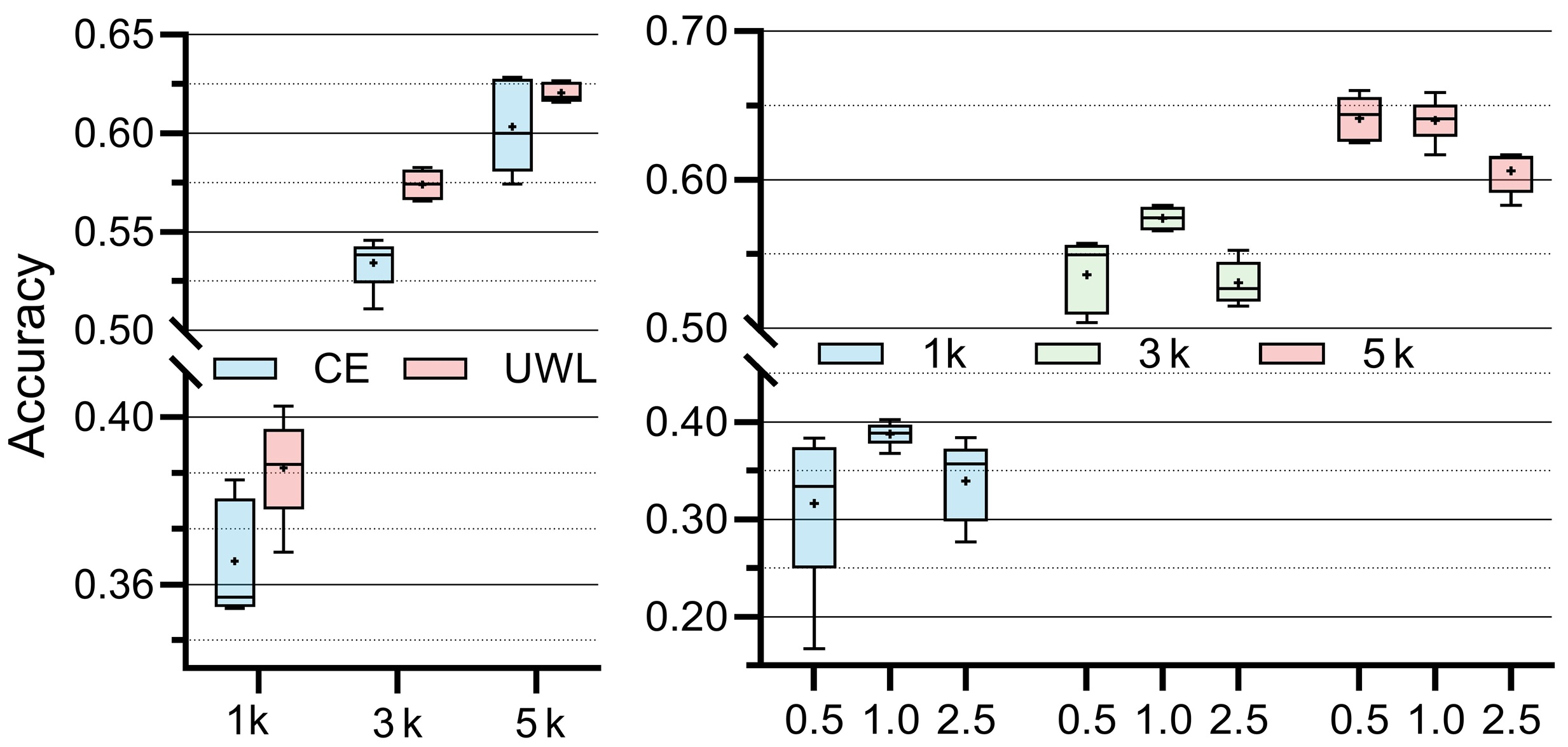}
\caption{Cervical cell classification accuracy results ($5$ independent trials) of different variations of the proposed method (the left plot) and under three different hyperparameter choices (the right plot) with the varying labeling costs of $1$K, $3$K and $5$K, respectively.}
\label{FigAblation}
\vspace{-4pt}
\end{figure}

We finally provide the ablation study results to show the reasonability of the method design.
We first look at the classification accuracy improvement results by the tailored updating rule, denoted by `UWL' in the left plot, Fig.~\ref{FigAblation}.
The variation, denoted by `CE', is that of our method removed by the tailored updating rule.
We conducted experiments under the labeling cost as $1$K, $3$K and $5$K, and produced the results by $5$ independent trials.
We can see that `UWL' works consistently better than `CE' in terms of both the classification accuracy and accuracy variance, suggesting that this updating rule is beneficial to the success of the proposed method.

We yet conducted the experiment for determining the value of $\alpha$ in the uncertainty normalization step.
We compared the classification accuracy results under three choices: $0.5$, $1.0$ and $2.5$ with $1$K, $3$K and $5$K labeling cost allowed.
Likewise, we produced results by $5$ independent trials.
We can see from the right plot in Fig.~\ref{FigAblation} that the proposed method produced the most accurate classification results with the smellest variance when $\alpha$ is set to $1.0$ in all labeling cost scenarios.
We therefore selected this value by default as our active labeling to produce the experimental results.

\section{Conclusion}

The quality of training dataset generally determines the learning performance of deep models.
Constructing a high representative training dataset, however, is non-trivial for cervical cell classification which is a crucial step in cervical cancer prevention.
We here presented an automatic construction method that prioritizes data for labeling per their contribution to the improvement of the classification performance by leveraging information of the classifier's uncertainty on the unlabeled cell images.
The experimental results confirm that the proposed method enables to construct a representative training dataset and in turn enhances the learning performance of cervical cell calssification.
This cost-effective method yet has a great potential to other applications where data-efficient learning is desired, opening the avenue for data efficient construction of the training dataset.

\end{document}